\documentclass[sensors,accept,moreauthors,pdftex,10pt,a4paper]{mdpi}
\firstpage{1}
\articlenumber{x}
\doinum{10.3390/------}
\pubvolume{xx}
\pubyear{2021}
\copyrightyear{2021}
\externaleditor{}
\history{}
\usepackage{bm}
\usepackage{multirow}
\usepackage{caption}
\usepackage{amsthm}
\usepackage{amssymb }
\usepackage{amsmath,tabularx}
\usepackage{bm}
\usepackage{bbm}
\usepackage{float}
\usepackage{mathrsfs}
\usepackage{amsfonts}
\usepackage{flushend}
\usepackage{subcaption}
\usepackage{graphicx}
\usepackage{algorithm, algpseudocode}
\usepackage[dvipsnames]{xcolor}
\def\MMK{\text{MAK-TD}}
\def\SR{\text{MAK-SR}}
\def\MSR{\text{MAK-TD/SR}}
\def\mS{\mathcal{S}}
\def\mA{\mathcal{A}}
\def\mP{\mathcal{P}}
\def\mR{\mathcal{R}}
\def\ua{_{a}}
\def\k{_{k}}
\def\nk{_{k+1}}
\def\j{^{j}}
\def\bt{\bm{\theta}}

\def\n{\bm{n}}
\def\m{\bm{m}}
\def\h{\bm{h}}

\def\I{\bm{I}}
\def\Q{\bm{Q}}

\def\K{\bm{K}}

\def\P{\bm{P}}
\def\F{\bm{F}}

\def\g{\bm{g}}

\def\Y{\bm{Y}}

\def\mZ{\mathcal{Z}}
\def\mbS{\mathbb{S}}
\def\u{\bm{\mu}}
\def\Sig{\bm{\Sigma}}

\def\s{\bm{s}}

\def\kpk{_{k|k-1}}

\def\M{\bm{M}}

\def\i{^{(i)}}

 \theoremstyle{mdpi}
 \newcounter{thm}
 \newcounter{ex}
 \newcounter{re}
 \theoremstyle{mdpidefinition}

\Title{Multi-Agent Reinforcement Learning via Adaptive Kalman Temporal Difference and Successor Representation}

\Author{Mohammad Salimibeni $^{1}$, Arash Mohammadi $^{1}*$, Parvin Malekzadeh $^{2}$, and Konstantinos N. Plataniotis $^{2}$}
\AuthorNames{Mohammad Salimibeni, Parvin Malekzadeh, Konstantinos N. Plataniotis, and Arash Mohammadi}

\address{%
$^{1}$ \quad Concordia Institute for Information System Engineering, Concordia University, Montreal, Canada; arashmoh@encs.concordia.ca\\
$^{2}$ \quad Department of Electrical and Computer Engineering, University of Toronto, Toronto, ON, Canada.}

\corres{Correspondence: arash.mohamad@concordia.ca; Tel.: +1-514-848-2712 ext. 2712}

\firstnote{Current address: 1455 De Maisonneuve Blvd. W. EV-009.187, Montreal, Quebec, Canada, HG-1M8}

\abstract{Development of distributed Multi-Agent Reinforcement Learning (MARL) algorithms has attracted an increasing surge of interest lately mainly due to the recent advancements of Deep Neural Networks (DNNs). Complex cooperative, competitive or mixed behavior among the agents in the multi-agent environments, make them more appealing to real world scenarios. Generally speaking, conventional Model-Based (MB) or Model-Free (MF) RL algorithms are not directly applicable to the MARL problems due to utilization of a fixed reward model for learning the underlying value function. While DNN-based solutions perform utterly well when a single agent is involved, such methods fail to fully generalize to the complexities of MARL problems. In other words, although recently developed approaches  based on DNNs for multi-agent environments have achieved superior performance, they  are still prone to overfiting, high sensitivity to parameter selection, and sample inefficiency. In this paper, an adaptive Kalman filter-based framework is introduced as an efficient alternative to address the aforementioned  problems. More specifically, the paper proposes the Multi-Agent Adaptive Kalman Temporal Difference  ($\MMK$) framework and its Successor Representation-based variant, referred to as the $\SR$. Intuitively speaking, the main objective is to capitalize on unique characteristics of Kalman Filtering (KF) such as uncertainty modeling and  online second order learning. The proposed $\MSR$ frameworks consider the continuous nature of the action-space that is associated with high dimensional multi-agent environments and exploit Kalman Temporal Difference (KTD) to address the parameter uncertainty. By leveraging the KTD framework, SR learning procedure is modeled into a filtering problem, where Radial Basis Function (RBF) estimators are used to  encode the continuous space into feature vectors. On the other hand, for learning localized reward functions, we resort to Multiple Model Adaptive Estimation (MMAE), as a remedy to deal with the lack of prior knowledge on the observation noise covariance and observation mapping function. The proposed $\MSR$ frameworks are evaluated via several experiments, which are implemented through the OpenAI Gym MARL benchmarks. In these experiments, different number of agents in cooperative, competitive and mixed (cooperative-competitive) scenarios are utilized. The experimental results illustrate  superior performance of the proposed $\MSR$ frameworks compared to their state-of-the-art counterparts.}


\begin{document}

\section{Introduction} \label{sec:Introduction}
Reinforcement Learning (RL), as a class of Machine Learning (ML) techniques, targets providing  human-level adaptive behavior by construction of an optimal control policy~\cite{Sutton}. Generally speaking, the main underlying objective is learning (via trial and error) from previous interactions of an autonomous agent and its surrounding environment. The optimal control (action) policy can be obtained via RL algorithms through the feedback that environment provides to the agent after each of its actions~\cite{salimiICASSP,Parvin:access, Spano, Seo, DrMing1, DrMing2}. Policy optimality can be reached via such an approach with the goal of increasing the reward over time. In most of the successful RL applications, e.g., Go and Poker games, robotics, and autonomous driving, typically, several autonomous agents are involved. This naturally falls within the context of Multi-Agent RL (MARL), which is a relatively long-established domain, however, it has recently been revitalized due to the advancements made in the single-agent RL approaches. In the MARL domain, which is the focus of this manuscript, multiple decision-making agents interact (cooperate and/or compete) in a shared environment to gain a common or a conflicting goal.

\vspace{.1in}
\noindent
\textbf{Literature Review:}
Traditionally, RL algorithms are classified as: (i) Model-Free (MF) approaches~\cite{Parvin:access, Turchetta2020, Jing2021} where sample trajectories are exploited for learning the value function, and; (ii) Model-Based (MB) techniques~\cite{Liu2020} where reward functions are estimated by leveraging search trees or dynamic programming~\cite{Bellman1954}. MF methods, generally, do not adapt quickly to local changes in the reward function.  On the other hand, MB techniques can adapt quickly to changes in the environment, but this comes with a high computational cost~\cite{Song2021, Sam, Vertes}. To address the above adaptation problems, Successor Representations (SR) approaches~\cite{Geerts2019, Machado2021} are proposed as an alternative RL category. The SR method provides the flexibility of the MB algorithm and has computational efficiency comparable to that of the MF algorithms. In SR-based methods, both the immediate reward expected to be received after each action and the discounted expected future state occupancy (which is called the SR) are learned. Afterwards, in each of the successor states, the value function is factorized into the SR and the immediate reward. This factorization only needs learning of the reward function for new tasks, allowing rapid policy evaluation when reward conditions are changed. In scenarios with a limited number of states,  the SR and the reward function (hence, the value function) associated with each state can be readily computed. Computation of the value function, however, is infeasible for MARL problems as in such scenarios we deal with a large number of continuous states~\cite{Moskovitz2021}. In other words, conventional approaches developed for single agent scenarios such as single-agent SR, Q-Learning, or policy gradient can not be directly adopted to MARL to compute the value function. The main problem here is that, typically, from a single agent's perspective, the environment tends to become unstable as each agent's policies change during the  training process. In the context of deep Q-learning~\cite{Hasselt2016}, this leads to stabilization issues as it is difficult to properly use the previous localized experiences. From the perspective of policy gradient, typically, observations demonstrate high variance in coordinating multiple agents.

To leverage SR-based solutions for MARL, value function approximation is unavoidable and one can use either linear or non-linear estimation approaches~\cite{Babu2012, Riedmiller2005}. In both categories, a set of adjustable parameters define the value of the approximated function. Non-linear function approximators, such as Deep Neural Networks (DNNs)~\cite{Riedmiller2005,Tang, Kim, Xie} have enabled application of RL methods to complex multi-agent scenarios. While DNN approaches like Deep Q-Networks (DQN)~\cite{Mnih2013} and Deep Deterministic Policy Gradient (DDPG)~\cite{Lillicrap2015} achieved superior results, they suffer from some major disadvantages including the overfitting problem; high sensitivity in choosing parameters; sample inefficiency, and; high number of episodes required for training the models. The linear function approximators, on the other hand, transform the approximation problem into a weight calculation problem in order to fuse several local estimators. Convergence can be examined when linear function approximators are utilized as they are better understood than their non-linear counterparts~\cite{Tsitsiklis1997, Bertsekas2004}. Cerebellar Model Articulation Controllers (CMACs)~\cite{Miller1990} and Radial Basis Functions (RBFs)~\cite{Haykin1994} are usually used as linear estimators in this context. It has been shown, however, that the function approximation process can be better represented via gradual-continuous transitions~\cite{Barreto2008}. Albeit the computation of the RBFs' parameters is usually based on prior knowledge of the problem at hand, these parameters can also be adapted leveraging observed transitions in order to improve the autonomy of the approach. In this context, cross entropy and gradient descent methods~\cite{Menache2005} can be utilized for the adaptation task. Stability of the gradient descent-based approach was later improved by exploiting a restrictive method in~\cite{Barreto2008}, which is adopted in this manuscript.

After verifying the value function's structure, to train the value function approximator, the following methodologies can be used: (i) Bootstrapping methods, e.g., Fixed-Point Kalman Filter (FPKF)~\cite{Choi2006}; (ii) Residual techniques such as Kalman Temporal Difference (KTD), and Gaussian Process Temporal Difference (GPTD)~\cite{Engel2005}, which is a special form of the KTD, and; (iii) Projected fixed-point methods such as Least Square Temporal Difference (LSTD)~\cite{Bradtke1996}. Among these methodologies, KTD~\cite{Geist2010} is a prominent technique as based on the selected structure, it provides both uncertainty and Minimum Mean Square Error (MMSE) approximation of the value function. In particular, uncertainty is beneficial for achieving higher sample efficiency. The KTD approach, however, requires prior knowledge of the filter's parameters (e.g., noise covariance of the process and measurement models), which are not readily available in realistic circumstances. Parameter estimation is a well-studied problem within the context of Kalman Filtering (KF), where several adaptive schemes are developed over the years including but not limited to Multiple Model Adaptive Estimation (MMAE) methods~\cite{Arash:TNSRE:2015, Arash:Sensors2016, Arash:TSP:2015} and, innovation-based adaptive schemes~\cite{Mehra1970}. When the system's mode is changing, the latter has the superiority to adapt faster and its efficiency was shown in~\cite{Assa2018}, where different suggested averaging and weighting patterns were compared. MMAE methods were already utilized in the RL problems, for instance, Reference~\cite{Kitao2017} proposed a multiple model KTD coupled with a model selection mechanism to address issues related to the parameter uncertainty. Existing multiple model methodologies are, however, not easily generalizable to the MARL problem.

In methods proposed in~\cite{Sam, Ma2018, Momennejad2017, Russek2017}, while the classical TD learning is coupled with DNNs, uncertainty of the value function and that of the SR are not studied. To deal with uncertainty, a good combination of exploitation and exploration should be used to prevent the agent's overconfidence about its knowledge to fully rely on exploitation. Alternatively, an agent can perform exploration over other possible actions, which might lead to improved results and a reduction in the uncertainty. Although, from computation points of view, it is intractable to find an optimal trade-off between exploitation and exploration, it has been represented that exploration can benefit from the uncertainty in two separate ways, i.e., through added randomness to the value function, and via shifting towards uncertain action selection~\cite{Sutton}. Consequently, the approximated value function's uncertainty, is a beneficial information for resolving the available conflict between exploration and exploitation~\cite{Sutton, Geist2013}.
It was shown in~\cite{Geist2013} that, the sensitivity of the framework to the parameters of the model can be diminished via uncertainty incorporation within the KTD method. Therefore, the required time and memory to find/learn the best model will be reduced compared to DNN-based methods~\cite{Sam, Ma2018, Momennejad2017, Russek2017}. The reduced sensitivity in setting the parameters enhances the reproducibility feature of a reliable approach, which leads to regeneration of more consistent outputs while running multiple learning epochs. Consequently the risk of getting unacceptable results in real scenarios will be decreased~\cite{Chan2020}. Geerts \textit{et al.}~\cite{Geerts2019} leveraged KTD framework to estimate the SR for problems with discrete state-spaces, however,  information related to uncertainty of the estimated SR is not considered in the action selection procedure. Malekzadeh \textit{et al.}~\cite{Parvin2021} applied the KTD approach within the SR learning context and used the uncertainty associated with the estimated SR for action estimation process, but only considered single agent environments.

\vspace{.1in}
\noindent
\textbf{Contributions:} The paper proposes Multi-Agent Adaptive Kalman Temporal Difference  ($\MMK$) framework and its SR-based variant, the Multi-Agent Adaptive Kalman Successor Representation ($\SR$) framework. The $\MSR$ frameworks consider the continuous nature of the action-space that is associated with high dimensional multi-agent environments and exploit KTD to address the parameter uncertainty. By leveraging the KTD framework, SR learning procedure is modeled into a filtering problem in this work. Intuitively speaking, the goal is to take advantage of the inherent benefits of the KF, i.e., online second order learning, uncertainty estimation and non-stationary handling. Afterwards, RBF-based estimation is utilized within the $\MSR$ frameworks in order for continuous states to be encoded into feature vectors and for the reward function to be projected as a linear function of the extracted feature vectors.
On the other hand, for learning localized reward functions, we resort to MMAE, as a remedy to deal with the lack of prior knowledge on observation noise covariance and observation mapping function. In summary, the paper makes the following key contributions:
\begin{itemize}
\item Within the MARL domain, the so-called $\MMK$ framework is proposed  as compensation for the information inadequacy about a key unknown filter's parameter, which is the measurement noise covariance. For learning the optimal policy and to simultaneously enhance sample efficiency of the proposed $\MMK$, an off-policy Q-learning approach is implemented.
\item  $\MMK$ is extended to $\SR$ by incorporation of the SR learning process into the filtering problem using KTD formulation for learned SR uncertainty approximation. Moreover, adopting KTD is beneficial to reduce the required memory/time to learn the SR while reducing model's sensitivity to parameters selection (i.e., more reliability) in comparison to DNN-based algorithms.
\item A coupled gradient descent and MMAE-based scheme is adopted for development of the $\SR$ framework to form a KF-based approximation of  the reward function. The ultimate objective is reducing sensitivity to having prior knowledge on the observation noise covariance and observation mapping function.
\item For establishing a trade off between exploration and exploitation, an innovative active learning mechanism is implemented to incorporate uncertainty of the value function obtained from the SR learning. Such a mechanism results in efficiently enhancing performance in terms of cumulative reward.
\end{itemize}
A multi-agent extension of the OpenAI gym benchmark, a two-dimensional world with continuous space~\cite{Mordatch2018} is utilized to simulate cooperative, competitive scenarios, and mix interaction settings. The proposed $\MSR$ frameworks are evaluated through a comprehensive set of experiments and simulations illustrating their superior performance compared to their counterparts.

The remainder of the paper is organized as follows: In Section~\ref{sec:PrbFor}, the basics of RL and MARL are briefly discussed. The proposed $\MMK$ framework is presented in Section~\ref{sec:MAMM-KTD}, and its SR-based variant, the $\SR$ framework, is introduced in Section~\ref{sec:AKF-SR}.  Experimental results based on multi-agent RL benchmark are presented in Section~\ref{sec:Sim}. Section~\ref{sec:con}, finally, concludes the paper.

\section{Problem Formulation} \label{sec:PrbFor}

To provide the background required for development of the proposed $\MSR$ frameworks, in this section, we present an overview of single agent and MARL techniques.

\subsection{Single-Agent Reinforcement Learning (RL)}
In conventional RL scenarios, typically, a single agent is placed in an unknown environment performing autonomous actions with the goal of maximizing its accumulated reward.  In such scenarios, the agent starts its interactions with the environment in an initial state denoted by $\s_0$ and continues to interact with the environment until reaching a pre-defined terminal state $\s_T$.  Action set $\mA$ is defined from which the agent can select potential actions following a constructed optimal policy. In other words, given its current state $\s\k \in \mS$, the single agent follows a policy denoted by $\pi\k$ and performs action $a\k \in \mA$ at time $k$. Following the agent's action, based on transition probability of $P(\s\nk|\s\k, a\k) \in \mP\ua$ it moves to a new state $\s\nk \in\mS$ receiving reward of $r\k \in \mR$. A discount factor $\gamma\in (0,1)$ is utilized to incorporate future rewards as such balancing the immediate rewards and future ones. In summary, a Markov Decision Process (MDP), denoted by $5$-tuple $\{\mS, \mA, \mP\ua, \mR, \gamma\}$, is, typically, used as the underlying mathematical model that governs the RL process. Therefore, the main objective is learning an optimal policy to map states into actions by maximizing the expected sum of discounted rewards, which is referred to as the optimal policy $\pi^*$~\cite{Sutton1998}. The optimal policy $\pi^*$ is, typically, obtained based on the following state-action value function
\begin{eqnarray}
 Q_{\pi}(\s, a) = \mathbb{E} \left\{\sum_{k=0}^{T}\gamma^k r\k |\s_0 = \s, a_0 = a, a\k = \pi(\s\k) \right\}.\label{Eq:Q}
\end{eqnarray}
Note that in Eq.~\eqref{Eq:Q},  $\mathbb{E}\{\cdot\}$ denotes the expectation operator. To perform an action at the learning stage, the current policy is utilized. Once convergence is reached, $a\k = \arg\max_{a\in\mA}Q_{\pi^*}(\s\k, a)$, which is the optimal policy, can be used by the agent to perform the required tasks. This completes a brief introduction to RL, next, the TD learning is reviewed as a building block of the proposed $\MSR$ frameworks.

\subsection{Off-Policy Temporal Difference (TD) Learning}
By taking an action and moving from one state to another, based on the Bellman equation and Bellman update scheme~\cite{Hutter2008}, the value function is gradually updated using sample transitions. This procedure is referred to as Temporal Difference (TD) update~\cite{Hutter2008}. There are two approaches to update policy namely ``on-policy learning'' or ``off-policy learning''. The former techniques use the current policy for action selection. For example, SARSA~\cite{Sutton1996,Xia2019} is an on-policy approach that optimizes the network as
\begin{eqnarray}
Q_{\pi}(\s\k, a\k) = Q_{\pi}(\s\k, a\k) +\alpha \Big(r\k +\gamma\,Q_{\pi}(\s\nk, a\nk)  - Q_{\pi}(\s\k, a\k) \Big),\label{Eq:TD}
\end{eqnarray}
where $\alpha$ denotes the learning rate and $Q_{\pi}(\s\k, a\k)$ is the state-action value function. In on-policy methods, by following a defined policy, selecting a new state becomes a non-optimal task. Additionally, this approach seems to be inefficient in sample selection since the value function is updated through the current policy instead of using the optimized one. In ``off-policy'' solutions, such as Q-learning~\cite{Watkins1992,Ge2019,Xia2019,Li2019}, the information received from previous policies is exploited to update the policy and reach a new one (exploitation). On the other hand, to properly explore new states, a stochastic  policy is usually chosen as the behavior policy (exploration). In brief, Q-learning is formed based on the Bellman optimal equation as follows
\begin{eqnarray}
Q_{\pi^*}(\s\k, a\k) = Q_{\pi^*}(\s\k, a\k) +\alpha \Big(r\k +\gamma \max_{a\in\mA}Q_{\pi^*}(\s\nk, a)  - Q_{\pi^*}(\s\k, a\k) \Big),\label{Eq:4}
\end{eqnarray}
where the optimal policy $\pi^*$ is used to form the state-action value function ${Q_{\pi^*}(\s\k, a\k)}$. The policy can be obtained via a greedy approach as follows
\begin{eqnarray}
V_{\pi^*}(\s) = \max_{a\in\mA}Q_{\pi^*}(\s\k, a).
\end{eqnarray}
Upon convergence, actions can be selected based on the optimal policy and not the behaviour policy as follows
\begin{eqnarray}
a\k = \arg\max_{a\in\mA}Q_{\pi^*}(\s\k, a).\label{Eq:6}
\end{eqnarray}
This completes our discussion on TD learning. In what follows, we  discuss the MARL approaches as well as value function approximation using the proposed algorithms in the multi-agent environments.

\subsection{Multi-Agent Setting}\label{Sec:MASs}
Within the context of MARL, we consider a scenario with $N$ agents, each with its localized observations, actions, and states. In other words, Agent $i$, for ($1 \leq i \leq N$), utilizes policy $\pi\i$, which is a function from the cartesian product of its localized action set $\mA\i$ and its localized observation set $\mZ\i$ to a real number within zero and one. We use superset $\mathbb{S} = \{\mS^{(1)}, \ldots, \mS^{(N)}\}$ to collectively represent all the localized states, $\mS\i$, for ($1 \leq i \leq N$). Likewise, supersets $\mathbb{A} = \{\mA^{(1)}, \ldots, \mA^{(N)}\}$ and $\mathbb{Z}  = \{\mZ^{(1)}, \ldots, \mZ^{(N)}\}$ are used to jointly represent all the localized actions, and local observations, respectively. Each agent makes localized decisions following the transition function $T : \mbS \times \mA^{(1)}\times, \ldots, \times \mA^{(N)} \rightarrow \mbS^{2}$. Consequently, an action is performed locally resulting in a new localized measurement and a localized reward $r\i : \mbS \times \mA\i \rightarrow {\mathbb{R}}$. The main objective of each agent is to maximize its localized expected return $R\i = \sum_{t = 0}^T \gamma^t (r\i)^t$ over a termination window of $T$ using a predefined discount factor of $\gamma$.

Traditional models like policy gradient or Q-Learning are not suitable for MARL scenarios~\cite{Lowe2017}, since the policy of an agent changes during the progress of the training, and the environment becomes non-stationary towards that specific agent's points of view. Consequently, most recently proposed platforms for multi-agent scenarios employ other strategies, where the agents' own observation (known as local information at the execution time) are exploited to learn optimal localized policies. Typically, such methods do not consider specific communication patterns between agents or any differentiable model of the environment's dynamics~\cite{Lowe2017}. Moreover, these models support different interactions between agents from cooperation to competition or their combination~\cite{Lowe2017,  Singh2019}.  In this context, an adaptation is made between the decentralized execution and centralized training  to be able to feed the policy training steps with more available data to speed up the process of finding the optimal policy.

\subsection{Multi-Agent Successor Representation (SR)}\label{Sec:SRs}
Within the context of SR, given an initial action $a\i$, and an initial state $\s\i$, the expected discounted future state occupancy of state ${\s'}\i$ is estimated based on the current policy $\pi\i$ as follows
\begin{eqnarray}
\M_{\pi\i}(\s\i,{\s'}\i,a\i)= \mathbb{E}\left[\sum_{k=0}^{T}\gamma^k \mathbbm{1}[\s\i\k={\s'}\i]|\s\i_0=\s\i, a\i_0=a\i\right],\label{Eq:7}
\end{eqnarray}
where $\mathbbm{1}\{\cdot\}=1$ if $\s\i\k={\s'}\i$, otherwise it is zero. The SR can be represented with a $N_{\s\i}\times N_{\s\i}$ matrix when the state-space is discrete.  The recursive approach used in Eq.~\eqref{Eq:TD}, can be leveraged to update SR as follows
\begin{eqnarray}
\lefteqn{\M_{\pi\i}^{\text{new}}(\s\i\k,{\s'}\i,a\i\k) = \M_{\pi\i}^{\text{old}}(\s\i\k,{\s'}\i,a\i\k) + \label{Eq:TD_SR} }\\
&&\alpha \Big(\mathbbm{1}[\s\i\k={\s'}\i] +\gamma\M_{\pi\i}(\s\i\nk,{\s'}\i,a\i\nk) - \M_{\pi\i}^{\text{old}}(\s\i\k,{\s'}\i,a\i\k) \Big).\nonumber
\end{eqnarray}
After computation (approximation) of the SR, its inner product with the estimated value of the immediate reward can be used to form the state-action value function based on Eq.~\eqref{Eq:Q}, i.e.,
\begin{eqnarray}
Q_{\pi\i}(\s\i\k, a\i\k) = \sum_{{\s'}\i\in \mS\i} \M(\s\i\k,{\s'}\i,a\i\k) R\i({\s'}\i,a\i\k). \label{nEq:Q_SR}
\end{eqnarray}
As a final note, it is worth mentioning an important characteristic of the SR-based approach, i.e., the state-action value function can be reconstructed based on the reward function. The developed MARL/MASR formulation presented here is used to develop the proposed $\MSR$ frameworks in the following sections.

\section{The $\MMK$ Framework} \label{sec:MAMM-KTD}
As stated previously, the $\MMK$ framework, is a Kalman-based off-policy learning solution for multi-agent networks. More specifically, by exploiting the TD approach represented in Eq.~\eqref{Eq:4}, the optimal value function associated with the $i^{th}$ agent, for ($1\leq i\leq N$), can be approximated from its one-step estimation as follows
\begin{eqnarray}
Q_{{\pi\i}^*}(\s\k\i, a\k\i) \approx r\k\i +\gamma \max_{{a\i}\in \mA} Q_{{\pi\i}^*}(\s\nk\i, a\i). \label{eq:Kml1}
\end{eqnarray}
By changing the variables' order, the reward at each time can be represented (modeled) as a noisy observation, i.e.,
\begin{equation}
r\k\i = Q_{{\pi\i}^*}(\s\k\i, a\k\i) -\gamma \max_{{a\i}\in \mA} Q_{\pi^*}(\s\nk\i, a\i) + v\k\i, \label{Eq:nn10}
\end{equation}
where $v\k$ is modeled as a zero-mean normal distribution with variance of $R\i$. By considering the local state-space of each agent, we use localized basis functions to approximate each agent's value function. Therefore, the following value function can be formed for Agent $i$, for ($1\leq i\leq N$),
\begin{eqnarray}
Q_{\pi\i}(\s_{k}\i, a_{k}\i) = \bm{\phi}(\s_{k}\i, a_{k}\i)^T \bt_{k}\i, \label{Eq:7}
\end{eqnarray}
where term $\bm{\phi\i}(\s\i, a\i)$ represents a vector of basis functions; $\pi\i$ is the policy associated with Agent $i$, and; finally, $\bt\i\k$ denotes the vector of the weights. Substituting Eq.~\eqref{Eq:7} in Eq.~\eqref{Eq:nn10} results in
\begin{equation}
r\k\i = \Big[\bm{\phi}(\s\k\i, a\k\i)^T- \gamma \max_{{a\i} \in\mA}\bm{\phi}(\s\nk\i, a\i)^T\Big]\bt\k\i+v\k\i, \label{Eq:11}
\end{equation}
which can be simplified into the following linear observation model
\begin{equation}
r\k\i = [\h\k\i]^T \bt\k\i+v\k\i,  \label{Eq:LOM}
\end{equation}
with
\begin{eqnarray}
\h\k\i = \bm{\phi}(\s\k\i, a\k\i) - \gamma \max_{{a\i} \in \mA} \bm{\phi}(\s\nk\i, a\i). \label{Eq:12n}
\end{eqnarray}
In other words, Eq.~\eqref{Eq:LOM} is the localized measurement (reward) of the $i^{\text{th}}$ agent, which is a linear model of the weight vector $\bm{\theta}\k\i$. For approximating localized weight $\bt\k\i$, first we leverage the observed reward, which is obtained by transferring from state $s\k\i$ to $s\nk\i$. Second, given that the noise variance of the measurement is not known a-priori, we exploit MMAE adaptation by representing it with $M$ different values (${R\j}\i$, for ($1 \leq j \leq M$). Consequently, a combination of $M$ KFs is used to estimate $\hat{\bt}\k\i$ based on each of its candidate values, i.e.,
\begin{eqnarray}
{\K\k\j}\i &=& \P_{(\bt,k|k-1)}\i \h\k\i \big({\h\k^T}\i \P_{(\bt, k|k-1)}\i \h\k\i +{R\j}\i \big)^{-1}\label{Eq:18}\\
\hat{\bt\k\j}\i &=& \hat{\bt}_{(k|k-1)}\i + {\K\k\j}\i \big(r\k\i - {\h\k^T}\i \hat{\bt}_{(k|k-1)}\i \big)\\
{\P\j_{\bt,k}}\i &=& \big(\I - {\K\k\j}\i {\h\k^T}\i \big) {\P^T_{(\bt, k|k-1)}}\i \big(\I - {\K\k\j}\i {\h\k^T}\i \big) +{\K\k\j}\i {R\j}\i {{\K\k\j}^T}\i,\label{Eq:23}
\end{eqnarray}
where superscript $j$ is used to refer to the $j^{\text{th}}$ matched KF, for which a specific value (${R\j}\i$) is assigned to model covariance of the observation model's noise process. The posterior distribution associated with each of the $M$ matched KFs is calculated based on its likelihood function. All the matched aposteriori distributions are then added together based on their corresponding weights to form the overall posterior distribution given by
\begin{eqnarray}
P\i(\bt\k|\Y\k) =\sum_{j=1}^{M}{\omega\j}\i P\i(\bt\k\i|\Y\k\i, {R\j}\i), \label{Eq:24}
\end{eqnarray}
where ${\omega\j}\i$ is the $j^{\text{th}}$ KF's normalized observation likelihood associated with the $i^{\text{th}}$ agent and is given by
\begin{equation}
{\omega\j}\i = P\i(r\k\i|\bt_{(k|k-1)}\i,R\j)=
c\i.e^{\big[\frac{-1}{2}\big(r\k\i-{\h\k^T}\i\hat{\bt}_{(k|k-1)}\i\big)^T\big({\h\k^T}\i\P_{(\bt,k|k-1)}\i\h\k\i+{R\j}\i\big)^{-1}}
{\big(r\k\i-{\h\k^T}\i\hat{\bt}_{(k|k-1)}\i\big)\big]},\label{Eq:25}
\end{equation}
where $c\i = 1/(\sum_{j=1}^{M} {w\j}\i)$. Exploiting Eq.~\eqref{Eq:24}, the weight and its error covariance are then updated as follows
\begin{eqnarray}
\hat{\bt}\k\i &=& \sum_{j=1}^{M}{\omega\j}\i \hat{\bt}{\k\j}\i \\
\P_{\bt,k}\i &=& \sum_{j=1}^{M}{\omega\j}\i \left({\P_{\bt,k}\j}\i + (\hat{\bt}{\j}\i-\hat{\bt}\i)(\hat{\bt}{\j}\i-\hat{\bt}\i)^T \right). \label{Eq:New28}
\end{eqnarray}
To finalize computation of $\hat{\bt}\k\i$ based on Eqs.~\eqref{Eq:LOM}-\eqref{Eq:New28}, localized measurement mapping function $\h\k\i$ is required. Since $\h\k\i$ is formed by the basis functions, its adaptation necessitates the adaptation of the basis functions. The vector of basis functions shown in Eq.~\eqref{Eq:7} is formed as follows
\begin{equation}
\bm{\phi}(\s\i\k) = \big[\phi_{1}(\s\i\k), \phi_{2}(\s\i\k),\ldots , \phi_{N_b-1}(\s\i\k), \phi_{N_b}(\s\i\k)\big]^T, \label{Eq:28n}
\end{equation}
where $N_b$ is the number of basis functions. Each basis function is represented by a RBF, which is defined by its mean and covariance parameters as follows
\begin{equation}
\phi_{n}(\s\i\k) = \exp\{\frac{-1}{2}(\s\i\k-\u\i_{n})^T{\bm{\Sigma}\i_{n}}^{-1}(\s\i\k-\u\i_{n})\}, \label{Eq:29}
\end{equation}
where $\u\i_{n}$ and $\bm{\Sigma}\i_{n}$ are the mean and covariance of $\phi_{n}(\s\i\k)$, for ($1 \leq n \leq N_b$). Generally speaking, the state-action feature vector can be represented as follows
\begin{eqnarray}
\bm{\phi}(\s\k\i, a\k\i) = [\phi_{1,a_{1}}(\s\i\k),\ldots \phi_{N_b,a_{1}}(\s\i\k), \phi_{1,a_{2}}(\s\i\k), \ldots \phi_{N_b,a_{D\i}}(\s\i\k)]^T,\label{Eq:AM1n}
\end{eqnarray}
where $\bm{\phi}(\cdot): \mA\i \times \mS \rightarrow \mathbb{R}^{N_b \times D\i}$, and $D\i$ denotes the number of actions associated with the $i^{\text{th}}$ agent. The state-action feature vector $\bm{\phi}(\s\i\k,a\i\k=a\i_d)$, for ($1 \leq d \leq D\i$) in Eq.~\eqref{Eq:AM1n} is considered to be generated from $\bm{\phi}(\s\i\k)$ by placing this state feature vector in the corresponding spot for action $a\i\k$ while the feature values for the rest of the actions are set to zero, i.e.,
\begin{equation}
\bm{\phi}(\s\i\k,a\i\k) = [0,\ldots 0, \phi_{1}(\s\i\k), \ldots, \phi_{N}(\s\i\k),0,\ldots 0,]^T. \label{Eq:30n}
\end{equation}
Due to the large number of parameters associated with the measurement mapping function, the multiple model approach seems to be inapplicable. Alternatively, Restricted Gradient Descent (RGD)~\cite{Barreto2008} is employed, where the goal is to minimize the following loss function
\begin{equation}
L\i\k = (\bm{\phi}^T(\s\i\k,a\k)\, \bm{\theta}\i\k - r\i\k)^2. \label{Eq:30}
\end{equation}
The gradient of the objective function with respect to the parameters of each basis function is then  calculated using the chain rule as follows
\begin{eqnarray}
\Delta\u\i &=& -\frac{\partial L\k\i}{\partial\u\i} =  -\frac{\partial L\k\i}{\partial Q_{{\pi^*}\i}}\frac{\partial Q_{{\pi^*}\i}}{\partial \bm{\phi\i}}\frac{\partial \bm{\phi\i}}{\partial\u\i}\\
\text{and }\Delta\Sig\i &=& -\frac{\partial\Sig\k\i}{\partial\u\i} =  -\frac{\partial L\k\i}{\partial Q_{{\pi^*}\i }}\frac{\partial Q_{{\pi^*}\i}}{\partial\bm{\phi\i}}\frac{\partial\bm{\phi\i}}{\partial\Sig\i},
\end{eqnarray}
where calculation of the partial derivations is done leveraging Eqs.~\eqref{Eq:7},~\eqref{Eq:29}, and \eqref{Eq:30}. Therefore, the mean and covariance of the RBFs can be adapted using the calculated partial derivative as follows
\begin{eqnarray}
 \u\i_{n} &=&  \u\i_{n} -2\lambda_{\u\i}{\left(L\i\k\right)}^{\frac{1}{2}} {\bm{\theta}\i\k}^T (\Sig\i_{n})^{-1} (\s\i\k -\u\i_{n})  \label{Eq:37mmk}\\
 \Sig\i_{n} &=& \Sig\i_{n} -  2\lambda_{\Sig\i}{\left(L\i\k\right)}^{\frac{1}{2}} {\bm{\theta\i}\k}^T  (\Sig\i_{n})^{-1}  \times (\s\i\k -\u\i_{n}) (\s\i\k -\u\i_{n})^T{\Sig\i_{n}}^{-1}, \label{Eq:38mmk}
\end{eqnarray}
where both $\lambda_{\u\i}$ and $\lambda_{\Sig\i}$ denote the adaptation rates. Based on~\cite{Barreto2008}, for the sake of stability, only one of the updates shown in Eqs.~\eqref{Eq:37mmk} and~\eqref{Eq:38mmk}, will be applied. To be more precise, when the size of the covariance is decreasing (i.e., ${L\i\k}^{\frac{1}{2}} ({\bm{\theta}\i\k}^T \bm{\phi}(\cdot))> 0$), the covariances of the RBFs are updated using Eq.~\eqref{Eq:38mmk}, otherwise their means are updated using Eqs.~\eqref{Eq:37mmk}. Using this approach, unlimited expansion of the RBF covariances is avoided.

One superiority that the proposed learning framework shows over other optimization-based techniques (e.g., gradient descent-based methods) is the calculation of the uncertainty for the weights $\P_{\bt,k}\i$, which is directly related to the uncertainty of the value function. This information can then be used at each step to select the actions, leading to the most reduction in the weights' uncertainty. Using the information form of the KF (information filter~\cite{AK2}), the information of the weights denoted by $\P_{\bt,k}\i$ is updated as follows
\begin{eqnarray}
{\P^{-1}_{\bt,k}}\i = {\P^{-1}_{(\bt,\kpk)}}\i +\h\k\i {R^{-1}}\i{\h\k^T}\i. \label{Eq:39}
\end{eqnarray}
In Eq.~\eqref{Eq:39}, the second element, i.e., $\h\k\i {R^{-1}}\i{\h\k^T}\i$, represents the information received from the measurement. The action is obtained by maximizing the information of the weights, i.e.,
\begin{eqnarray}
a\k\i &=& \arg\max_a \Big(\h\k\i (\s\k\i, a\i){R^{-1}}\i{\h\k^T}\i(\s\k\i, a\i) \Big)\nonumber\\
&=&\arg\max_a \Big(\h\k\i (\s\k\i,a\i){\h^T\k}\i(\s\k\i, a\i) \Big).\label{Eq:40}
\end{eqnarray}
The second equality in Eq.~\eqref{Eq:40} is constructed as $R\i$ is a scalar. The projected behaviour policy in Eq.~\eqref{Eq:40} is different from that of Reference~\cite{Geist2010}, where a random policy was proposed, which favored actions with less certainty of the value function. Although reducing the value function's uncertainty through action selection is an intelligent approach, it is less efficient in sample selection due to the random nature of such policies. Algorithm~\ref{algo:1}, briefly represents the $\MMK$ framework proposed in this work.
\begin{algorithm*}[t!]
\setstretch{1.75}
\caption{\textproc{The Proposed $\MMK$ Framework}}
\label{algo:1}
\begin{algorithmic}[1]
\State \textbf{Learning Phase:}
\State Set $\bt_0, \P_{\bt,0},\F,\u_{n,i_d}, \Sig_{n,i_d}$ for $n={1,2,\hdots,N}$ and $i_d={1,2,\hdots,D}$
\State  \textbf{Repeat} (for each episode):
\State \quad \quad Initialize $\s\k$
\State \quad \quad \textbf{Repeat} (for each agent $i$):
\State \quad \quad \quad \textbf{While} $\s\k\i \neq \s_T$ \textbf{do}:
\State \quad \quad \quad \quad $a\k\i =\arg\max\limits_a \Big(\h\k\i  (\s\k\i ,a\i){\h\k ^T}\i(\s\k\i , a\i) \Big)$
\State \quad \quad \quad \quad Take action $a\k\i $ , observe $\s\nk\i , r\k\i $
\State \quad \quad \quad \quad Calculate $\bm{\phi\i}(\s\i,a\i)$ via Eqs.~\eqref{Eq:28n} and~\eqref{Eq:29}
\State \quad \quad \quad \quad $\h\k\i (\s\k\i ,a\k\i )=\bm{\phi\i}(\s\k\i ,a\k\i )-\gamma\arg\max\limits_a \bm{\phi\i}(\s\nk\i ,a\i)$
\State \quad \quad \quad \quad  $\hat{\bt}_{(k|k-1)}\i=\F\i {\hat{\bt}\k}\i$
\State \quad \quad \quad \quad $\P_{(\bt,k|k-1)}\i=\F\i \P_{\bt,k-1}\i {\F^T}\i +\Q\i$
\State \quad \quad \quad \quad \textbf{for} $j=1:M$ \textbf{do}:
\State \quad \quad \quad \quad \quad \quad ${\bm{k}\k\j}\i=\P_{(\bt,k|k-1)}\i \h\k\i ({\h\k^T}\i \P_{(\bt,k|k-1)}\i\h\k\i+{R\j}\i)^{-1}$
\State \quad \quad \quad \quad \quad \quad $\hat{\bt}{\k\j}\i=\hat{\bt}_{(\bt,k|k-1)}\i+{\bm{k}\k\j}\i (r\k\j-\h{\k^T}\i \hat{\bt}_{(k|k-1)}\i)$
\State  \quad \quad \quad \quad \quad \quad { $\P_{\bt,k}\i=(\I-{\bm{K}\k\j}\i {\h\k^T}\i)\P_{(\bt,k|k-1)}\i (\I-{\bm{K}\k\j}\i {\h\k^T}\i)^T
+{\bm{K}\k\j}\i R\j {{\bm{K}\k\j}^T}\i$}
\State \quad \quad \quad \quad \textbf{end for}
\State \quad \quad \quad \quad { Compute the value of $c$ and ${w\j}\i$ by using $\sum_{j=1}^{M} {w\j}\i=1$ and Eq.~\eqref{Eq:25}}
\State \quad \quad \quad \quad $\hat{\bt}\k\i=\sum_{j=1}^{M} {w\j}\i \hat{\bt}{\k\j}\i $
\State \quad \quad \quad \quad $\P_{\bt\k}\i = \sum_{j=1}^{M}{\omega\j}\i\left({\P_{\bt,\k}\j}\i +(\hat{\bt}{\j}\i-\hat{\bt}\i)(\hat{\bt}{\j}\i-\hat{\bt}\i)^T \right)$
\State \quad \quad \quad \quad \textbf{RBFs Parameters Update:}
\State \quad \quad \quad \quad $L\i\k = (\bm{\phi}^T(\s\i\k,a\k)\, \bm{\theta}\i\k - r\i\k)^2$
\State \quad \quad \quad \quad \textbf{if} ${L\i\k}^{\frac{1}{2}} ({\bm{\theta}\i\k}^T \bm{\phi}(\cdot))> 0$ \textbf{then}:
\State \quad \quad \quad \quad \quad Update $\Sig_{n,a_d}$ via Eq.~\eqref{Eq:37mmk}
\State \quad \quad \quad \quad \textbf{else:}
\State \quad \quad \quad \quad \quad Update $\u_{n,a_d}$ via Eq.~\eqref{Eq:38mmk}
\State \quad \quad \quad \quad \textbf{end if}
\State \quad \quad \quad \textbf{end while}
\State \textbf{Testing Phase:}
\State \textbf{Repeat} (for each trial episode):
\State \quad \quad \textbf{While} $\s\k \neq \s_T$ \textbf{do}:
\State \quad \quad \quad \textbf{Repeat} (for each agent):
\State \quad \quad \quad \quad $a\k=\arg \max\limits_{a}\bm{\phi}(\s\k, a)^T \bt\k$
\State \quad \quad \quad \quad Take action $a\k$, and observe $\s\nk,r\k$
\State \quad \quad \quad \quad Calculate Loss $S_k$ for all agents
\State \quad \quad \textbf{End While}
\end{algorithmic}
\end{algorithm*}
\section{The $\SR$ Framework} \label{sec:AKF-SR}
In the previous section, $\MMK$ framework is proposed, which is a MM Kalman-based off-policy learning solution for multi-agent networks. To learn the value function, a fixed model for the reward function is considered, which could restrict its application to more complex MARL problems. SR-based algorithms are appealing solutions to tackle this issue where the focus is instead on learning the immediate reward and the SR, which is the expected discounted future state occupancy. In the existing SR-based approaches that use standard temporal difference methods the uncertainty about the approximated SR, is not captured.  In order to address this issue, we extend the  $\MMK$ framework  and design its SR-based variant in this section. In other words, $\MMK$ is extended to $\SR$ by incorporation of the SR learning procedure into the filtering problem using KTD formulation to estimate uncertainty of the learned SR. Moreover, by applying KTD, we benefit from the decrease in memory and time spent for the SR learning and also sensitivity of the framework’s performance to its parameters (i.e., more reliable) when compared to DNN-based algorithms.

Exact computation of the SR and the reward function is, typically, not possible within the multi-agent settings as we are dealing with a large number of continuous states. Therefore, we follow the approach developed in Section~\ref{sec:MAMM-KTD} and approximate the SR and the reward function via basis functions. For the state-action feature vector $\bm{\phi}(\s\i,a\i)$, a feature-based SR, which encodes the expected occupancy of the features, is defined as follows
\begin{equation}
\M_{\pi\i}(\s\i,:,a\i)= \mathbb{E}
\left[\sum_{k=0}^{T}\gamma^k \bm{\phi}(\s\i\k,a\i\k)|\s\i_0=\s\i, a\i_0=a\i\right]. \label{Eq:Sr_appr}
\end{equation}
We consider that the immediate reward function for pair $(\s\i,a\i)$ can be linearly factorized as
\begin{eqnarray}
r\i(\s\i\k,a\i\k)\approx\bm{\phi}(\s\i\k,a\i\k)^T \bt\i\k, \label{Eq:reward}
\end{eqnarray}
where $\bt\i\k$ is the reward weight vector. The state-action value function (Eq.~\eqref{nEq:Q_SR}), therefore, can be computed as follows
\begin{eqnarray}
Q(\s\i\k,a\i\k)={\bt\k\i}^T \M(\s\i\k,:,a\i\k).\label{Eq:Q_estimation}
\end{eqnarray}
The SR matrix $\M(\s\i\k,:,a\i\k)$ can be approximated as a linear function of the same feature vector as follows
\begin{eqnarray}
\M_{\pi\i}(\s\i\k,:,a\i\k)\approx \M\k\, \bm{\phi}(\s\i\k,a\i\k). \label{Eq:linear_SR}
\end{eqnarray}
The TD learning of the SR then can be performed as follows
\begin{equation}
\M_{\pi\i}^{\text{new}}(\s\i\k,:,a\i\k)=\M_{\pi\i}^{\text{old}}(\s\i\k,:,a\i\k)+\alpha\big(\bm{\phi\i}(\s\i\k,a\i\k)+ \gamma \M_{\pi\i}(\s\i\nk,:,a\i\nk)-\M_{\pi\i}^{\text{old}}(\s\i\k,:,a\i\k)\big).
\label{Eq:TD_SR2}
\end{equation}
By defining the estimation structure of the SR and reward function, a suitable method must be selected to learn (approximate) the weight vector of the reward $\bt\i$ and the weight matrix of the SR $\M$ for Agent $i$. The proposed multi-agent $\SR$ algorithm contains two main components, i.e., KTD-based weight SR learning and radial basis function update. For the latter, we apply the method developed in Section~\ref{sec:MAMM-KTD} to approximate the vector of basis functions via representing each of them as a RBF. The gradient of the loss function~\eqref{Eq:30}, with respect to the parameters of the RBFs is calculated using the chain rule for the mean and covariance of RBFs using~\eqref{Eq:37mmk} and~\eqref{Eq:38mmk}.

For KTD-based weight SR learning, the SR can be obtained from its one-step approximation using the TD method of Eq.~\eqref{Eq:TD_SR2}. In this regard, the state-action feature vector at time step $k$ can be considered as a noisy measurement from the system as follows
\begin{equation}
\hat{\bm{\phi}}(\s\i\k,a\i\k) =\M^{\text{new}}(\s\i\k, :,a\i\k) - \gamma \M(\s\i\nk, :,a\i\nk) + \n\i\k, \label{Eq:10}
\end{equation}
where $\n\i\k$ follows a zero-mean normal distribution with covariance of $\bm{R}\i_{\M}$. Considering Eqs.~\eqref{Eq:linear_SR} and~\eqref{Eq:10} together, the feature vector $\bm{\phi}(\s\i\k,a\i\k)$ can be approximated as
\begin{equation}
\hat{\bm{\phi}}(\s\i\k,a\i\k) = \M\k \underbrace{\Big[\bm{\phi}(\s\i\k,a\i\k)- \gamma \bm{\phi}(\s\i\nk,a\i\nk)\Big]}_{\g\i\k}+\n\i\k. \label{Eq:11}
\end{equation}s
Matrix $\M\k$ is then mapped to a column vector $\m\i\k$ by concatenating its columns. Using the vec-trick characteristic of Kronecker product denoted by $\otimes$, then we can rewrite Eq.~\eqref{Eq:11}  as follows
\begin{eqnarray}
\hat{\bm{\phi}}(\s\i\k,a\i\k) = ({\g\i}^T\k \otimes \I)\m\i\k + \n\i\k, \label{Eq:Kronecker}
\end{eqnarray}
where $\I$ represents an identity matrix of appropriate dimension. More specifically,  Eq.~\eqref{Eq:Kronecker} is used to represent the localized measurements ($\bm{\phi}(\s\i\k,a\i\k)$) linearly based on vector $\m\i\k$, which requires estimation. Therefore, we use the following linear state model
\begin{eqnarray}
\m\i\nk = \m\i\k + \u\i\k, \label{Eq:SSpace}
\end{eqnarray}
to complete the required state-space representation for KF-based implementation. The noise associated with the state model (Eq.~\eqref{Eq:SSpace}), i.e.,  $\u\i\k$, follows a zero-mean normal distribution with covariance of $\Q_{\M}$. Via implementing the KF's recursive equations, we use the new localized observations to estimate $\m\i\k$ and its corresponding covariance matrix $\P\i_{\m\i,k}$. After this step, vector $\m\i\k$ is reshaped to form a ($L\times L$) matrix in order to reconstruct Matrix $\M\k$. Eq.~\eqref{Eq:Q_estimation} is finally used to form the  state-action value function for associated with ($\s\i\k,a\i\k$). Algorithm~\ref{algo:2} summarizes the proposed $\SR$ framework.
\begin{algorithm*}[!t]
\setstretch{1.75}
\caption{\textproc{The Proposed $\SR$ Framework}}
\label{algo:2}
\begin{algorithmic}[1]
\State \textbf{Learning Phase:}
\State \textbf{Initialize:} $\bt_0, \P_{\bt,0},\m_0, \P_{\M,0}, {\u_{n}, \text{and}\, \Sig_{n}}$ for $n={1,2,\hdots,N}$
\State \textbf{Parameters:} $ \Q_{\bt}, \Q_{\M}, \lambda_{\u},  \lambda_{\Sig}, \text{and}\, \{R\j_{\bt}, R\j_{\M}\}$ for $j={1,2,\hdots,M}$
\State \textbf{Repeat} (for each episode):
\State \quad Initialize $\s\k$
\State \quad \textbf{Repeat} (for each agent $i$):
\State \quad \quad \textbf{While} $\s\k\i \neq \s_T$ \textbf{do}:
\State \quad \quad \quad {Reshape $\m\k$ into $L \times L$ to construct 2-D matrix $\M\k$.}
\State \quad \quad \quad $a\k\i=\arg\max\limits_a \Big(\g\k\i (\s\k\i,a){\g\k\i}^T(\s\k\i, a\i) \Big)$
\State \quad \quad \quad Take action $a\k\i$ , observe $\s\nk\i$ and $r\k\i$.
\State \quad \quad \quad Calculate $\bm{\phi}(\s\k\i,a\k\i)$ via Eqs.~\eqref{Eq:29} and~\eqref{Eq:30n}.
\State \quad \quad \quad {\textbf{Update reward weights vector:} Perform MMAE to update $\bt\k\i$.}
\State \quad \quad \quad {\textbf{Update SR weights vector:} Perform KF on Eqs.~\eqref{Eq:Kronecker} and~\eqref{Eq:SSpace} to update $\m\k\i$.}
\State \quad \quad \quad {\textbf{Update RBFs parameters:} Perform RGD on the loss function $L\k$ to update $\Sig_{n}$ and $\u_{n}$.}
\State \quad \quad \textbf{end while}
\end{algorithmic}
\end{algorithm*}
It is worth mentioning that, unlike the  DNN-based networks for multi-agent scenarios, the proposed multiple-model frameworks require far less memory due to their sequential data processing nature. In other words, storing the whole episodes' information for all the agents is not needed as the last measured data (assuming one-step Markov decision process) can be leveraged given the sequential nature of the incorporated filters. Finally, it should be noted that the proposed  $\SR$  and $\MMK$ frameworks are designed for systems with a finite number of actions. . One direction for future research is to consider extending the proposed $\SR$ framework to applications where the to  action-space is infinite-dimensional. This might occur in continuous control problems~\cite{Li2019, Zhang:2021} where number of possible actions at each state is infinite.

\section{Experimental Results}  \label{sec:Sim}
\begin{figure}[t!]
\centering
\includegraphics[scale=0.3]{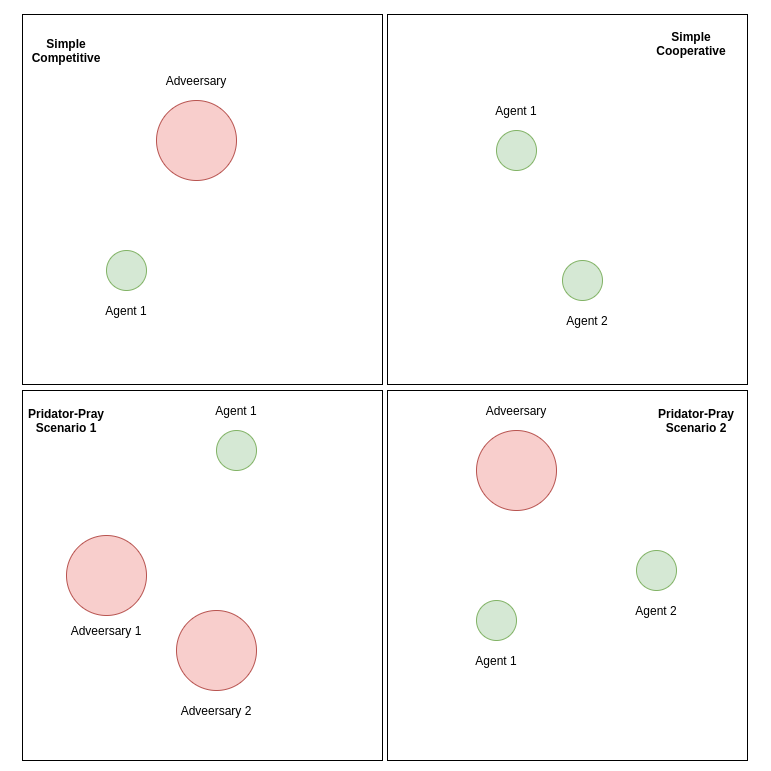}
\caption{\small Different multi-agent scenarios implemented within the OpenAI gym.}\label{Fig:2}
\end{figure}
The performances of the proposed $\SR$ and $\MMK$  frameworks are evaluated in this section, where a multi-agent extension of the OpenAI gym benchmark is utilized. Fig.~\ref{Fig:2} illustrates snapshots of the environment utilized for evaluation of the proposed approaches. More specifically, a two-dimensional world is implemented to simulate competitive, cooperative, and/or mix interaction scenarios~\cite{Mordatch2018}. The utilized benchmark is currently one of the most standard environments to test different multi-agent algorithms, where time, discrete action space, and continuous observations are the basics of the environment. Such a multi-agent environment is a natural curriculum in that the environment difficulty is determined based on the skills of the agents cooperating or competing. The environment does not have a stable equilibrium, therefore, allowing the participating agents to become smarter irrespective of their intelligence level. In each step,  the implemented environment provides observations and rewards once the agents performed their actions.

In what follows, we discuss different multi-agent environments exploited in this work as well as the experimental assumptions considered during testing of the proposed methods. Finally, the results of the experiments will be represented and explained.

\subsection{Environments}  \label{sec:Environments}
In the represented multi-agent environments,  we do not impose any assumption or requirement on having identical observations or action spaces for the agents. Furthermore, agents are not restricted to follow the same policy $\pi$ while playing the game. In the  environments, a different number of agents and possible landmarks can be placed to establish different interactions such as  cooperative, competitive, or mixed strategies. The strategy in each environment is to keep the agents in the game as long as possible. Each test can be fully cooperative when agents communicate to maximize a shared return, or can be fully competitive when the agents compete to achieve different goals. The mixed scenario for the predator-prey environments (a variant of the classical predator-prey) is defined in a way that a group of slower agents must cooperate against another group of faster agents to maximize their returned reward.  Each agent takes a step by choosing one of five available actions, i.e., no movement, left, right, up, and down, transiting to a new state, and receiving a reward from the environment. Moreover, each agent will receive a list of observations in each state, which  contains the agent's position and velocity, relative positions of landmarks (if available), and its relative position to other agents in the environment. That is how an agent knows the position and general status of the  agents (friends and adversaries), enabling the decision making process of that agent. As shown in Fig.~\ref{Fig:2}, each environment has its own margins. An agent that leaves the area  will be punished by $-50$ points, the game will be reset, and a random configuration will be initiated to start the next state, which begins immediately. The red agents play the predator role and receive $+100$ points intercepting (hunting) a prey (small green agents). The green agents that are faster than red agents (predators) will receive $-100$ points by each interception with the red ones. As their job is to follow the prey, the predators will be punished proportionally to their distance to the prey (green agents). In contrast, the opposite will happen to the green agents as they keep the maximum distance from the predators. The proposed $\MSR$ frameworks are evaluated against DQN~\cite{Mnih2013}, DDPG~\cite{Lillicrap2015}, and MADDPG~\cite{Lowe2017}.  We evaluate the algorithms in terms of  loss, returned discounted reward, and the number of collisions between agents.

\subsection{Experimental Assumptions} \label{sec:Environments}
\begin{table*}[t]
\vspace{-.1in}
\caption{\small  Total loss averaged across all the episodes and for all the four implemented scenarios.}\label{Table:table1}
\centering
\begin{tabular}{|c|p{1.6cm}|p{1.6cm}|p{1.2cm}|p{0.9cm}|p{0.7cm}|}
\hline
\multirow{2}[2]{*}{\textbf{\ Environment}} & \multirow{2}[2]{*}{\textbf{$\SR$}} & \multirow{2}[2]{*}{\textbf{$\MMK$}} &  \multirow{2}[2]{*}{\textbf{MADDPG}} &  \multirow{2}[2]{*}{\textbf{DDPG}}&  \multirow{2}[2]{*}{\textbf{DQN}} \\
&&&&\\
\hline
\multirow{1}[1]{*}{\textbf{Simple Cooperation}} & 8.93 & 2.4088 &  9649.84 &  10561.16 &  10.93 \\
\hline
\multirow{1}[1]{*}{\textbf{Simple Competition}} & 0.43 & 4.9301 &  10158.18 &  10710.37 &  107.39 \\
\hline
\multirow{1}[1]{*}{\textbf{Predator-Prey 1v2}} & 0.005 & 1.9374 &  6816.34 &  6884.33 &  8.21 \\
\hline
\multirow{1}[1]{*}{\textbf{Predator-Prey 2v1}} & 8.87 & 1.2421 &  7390.18 &  6882.2 &  10.24 \\
\hline
\end{tabular}
\end{table*}
\begin{table*}[t]
\caption{\small Total received reward by the agents averaged for all the four implemented scenarios.}\label{Table:table3}
\centering
\begin{tabular}{|c|p{1.6cm}|p{1.6cm}|p{1.2cm}|p{0.9cm}|p{0.9cm}|}
\hline
\multirow{2}[2]{*}{\textbf{\ Environment}} & \multirow{2}[2]{*}{\textbf{$\SR$}} & \multirow{2}[2]{*}{\textbf{$\MMK$}} &  \multirow{2}[2]{*}{\textbf{MADDPG}} &  \multirow{2}[2]{*}{\textbf{DDPG}}&  \multirow{2}[2]{*}{\textbf{DQN}} \\
&&&&\\
\hline
\multirow{1}[1]{*}{\textbf{Simple Cooperation}} & -16.0113 & -23.0113 &  -69.28 &  -66.29 &  -39.96 \\
\hline
\multirow{1}[1]{*}{\textbf{Simple Competition}} & -0.778 & -13.358 &  -63.30 &  -61.34 &  -14.49 \\
\hline
\multirow{1}[1]{*}{\textbf{Predator-Prey 1v2}} & -0.0916 & -13.432 &  -46.17 &  -20.53 &  -23.451 \\
\hline
\multirow{1}[1]{*}{\textbf{Predator-Prey 2v1}}  & -0.081 & -17.0058 &  -55.69 & -49.41  &  -44.32 \\
\hline
\end{tabular}
\end{table*}
\begin{table*}[t]
\caption{\footnotesize Average steps taken by agents per episode for all the environments based on the implemented platforms.}\label{Table:table4}
\centering
\begin{tabular}{|c|p{1.6cm}|p{1.6cm}|p{1.2cm}|p{0.9cm}|p{0.7cm}|}
\hline
\multirow{2}[2]{*}{\textbf{\ Environment}} & \multirow{2}[2]{*}{\textbf{$\SR$}} & \multirow{2}[2]{*}{\textbf{$\MMK$}} &  \multirow{2}[2]{*}{\textbf{MADDPG}} &  \multirow{2}[2]{*}{\textbf{DDPG}}&  \multirow{2}[2]{*}{\textbf{DQN}} \\
&&&&\\
\hline
\multirow{1}[1]{*}{\textbf{Simple Cooperation}} & 14.03 & 12.064 &  7.377 &  7.369 &  15.142 \\
\hline
\multirow{1}[1]{*}{\textbf{Simple Competition}}  & 17.59 & 17.48 &  7.36 &  7.18 &  11.98 \\
\hline
\multirow{1}[1]{*}{\textbf{Predator-Prey 1v2}} & 14.78 & 12.36 &  6.21 &  7.69 &  10.02 \\
\hline
\multirow{1}[1]{*}{\textbf{Predator-Prey 2v1}} & 9.94 & 9.773 &  6.25 &  7.12 &  8.46 \\
\hline
\end{tabular}
\end{table*}
In the proposed frameworks, we exploit related RBFs based on the different agents' sizes of observations and a bias parameter. The size of the observation vector at each local agent (localized observation vector), which represents the number of global and local measurements available locally, varies across different scenarios based on the type and the number of agents present/active in the environment. Irrespective of  size of the localized observation vectors, the size of the localized feature vectors, which represents the available five actions, is considered to be $50$. Mean and covariance of the RBFs are initialized randomly for all the agents in all the environments. For example, consider a Predator-Prey scenario with $2$ preys optimizing their actions against one predator. In this toy-example (discussed for clarification purposes), considering $9$ RBFs together with localized observation vectors of size $12$ for the predator and $10$ for the preys, the mean vector associated with the predator and the preys are of dimensions $9\times 12$ and $9\times 10$, respectively. Consequently, for this Predator-Prey scenario, $\bm{\mu}$,  which is initialized randomly contains three agents with random values with the mean size $((9,12), (9,10), (9,10))$ and the covariance, $\Sig = (\I_{12}, \I_{10}, \I_{10})$ where $\I_{12}$ and $\I_{10}$ are the identity matrices of size ($12\times 12$) and ($10\times 10$),  respectively. Based on Eq.~\eqref{Eq:30n}, the vector of basis function is represented as follows
\begin{eqnarray}
\bm{\phi}(\s\k,a\k=-2) &=& [0, \ldots ,0, 0, \ldots ,0, 1,\phi_{1,a_d},\ldots \phi_{9,a_d},
0, \ldots ,0, 0, \ldots ,0 ]^T, \\
\bm{\phi}(\s\k,a\k=-1) &=& [0, \ldots ,0, 0, \ldots ,0, 0, \ldots,0,
1,\phi_{1,a_d},\ldots \phi_{9,a_d},0, \ldots ,0 ]^T, \\
\bm{\phi}(\s\k,a\k=0) &=& [0, \ldots ,0, 0, \ldots ,0, 0, \ldots ,0, 0, \ldots ,0,
1,\phi_{1,a_d},\ldots \phi_{9,a_d}]^T, \\
\bm{\phi}(\s\k,a\k=+1) &=& [0, \ldots ,0, 1,\phi_{1,a_d},\ldots ,\phi_{9,a_d}, 0, \ldots 0,
0,\ldots, 0, 0, \ldots ,0]^T \\
\text{and} \quad\quad \bm{\phi}(\s\k,a\k=+2) &=& [1,\phi_{1,a_d},\ldots ,\phi_{9,a_d}, 0, \ldots ,0,0,\ldots, 0,
0, \ldots 0, 0, \ldots 0]^T,
\end{eqnarray}
where $\phi_{l,a_d}$ is calculated based on Eq.~\eqref{Eq:AM1n} for ($l \in \{1,2,\dots,9\}$. , $\gamma$, In all the scenarios, the time step chosen to be $10$ milliseconds and the discount factor is  $0.95$. The transition matrix is initiated to $\F = \I_{50}$, and  for the process noise covariance, a small value of $\Q\k = 10^{-7}\I_{50}$ is considered. The covariance matrix associated with the noise of the measurement model is selected from the following set
\begin{eqnarray}
R\i \in \{0.01, 0.1, 0.5, 1, 5, 10, 50, 100\}.
\end{eqnarray}
For initializing the weights, we sample from a zero mean Gaussian initialization distribution $\mathcal{N}(\bt_0, \P_{\bt,0})$, where $\bt_0 = \bm{0}_{50}$ and $\P_{\bt,0} = 10\I_{50}$. By considering the aforementioned initial parameters, each experiment is initiated randomly and consists of $1,000$ learning episodes together with $1,000$ test episodes. Given small number of available learning episodes, the proposed $\MSR$ frameworks outperformed their counterparts across different metrics including sample efficiency, cumulative reward, cumulative steps, and speed of the value function convergence.
\begin{figure}[t!]
\centering
\includegraphics[scale=0.45]{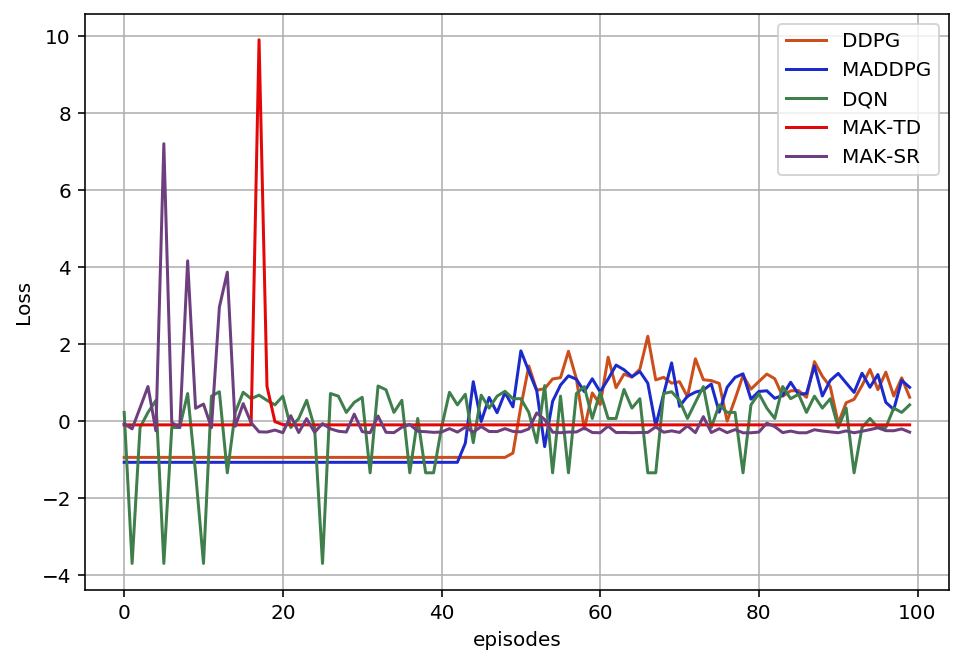}
\includegraphics[scale=0.45]{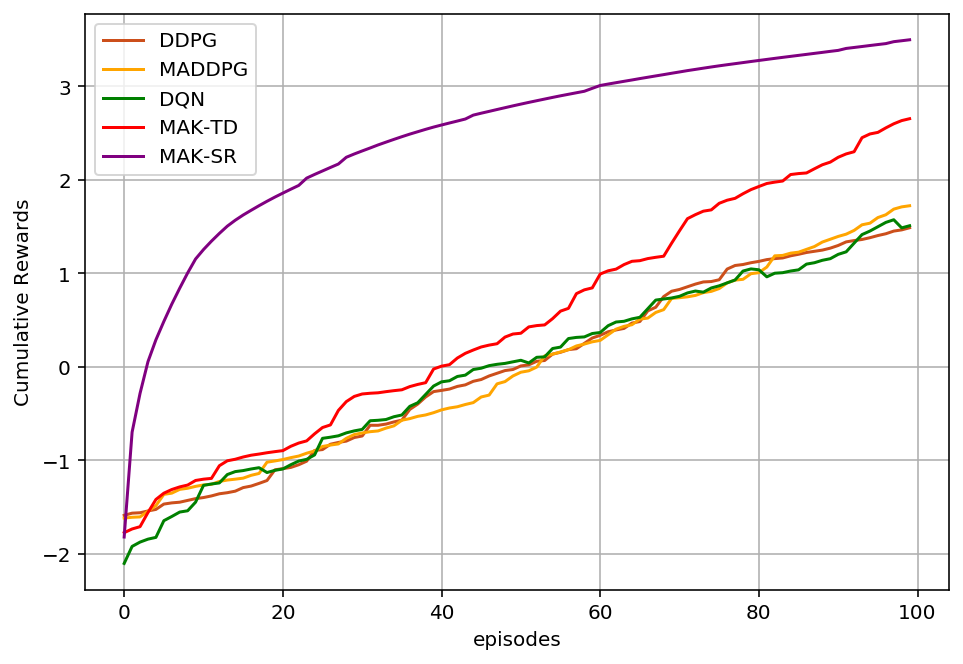}
\caption{\footnotesize The Predator-Prey environment: (a) Loss. (b) Received rewards.}\label{Fig:fig3}
\end{figure}
\begin{figure}[t!]
\centering
  \begin{subfigure}{6cm}
    \centering
    \includegraphics[width=6.3cm]{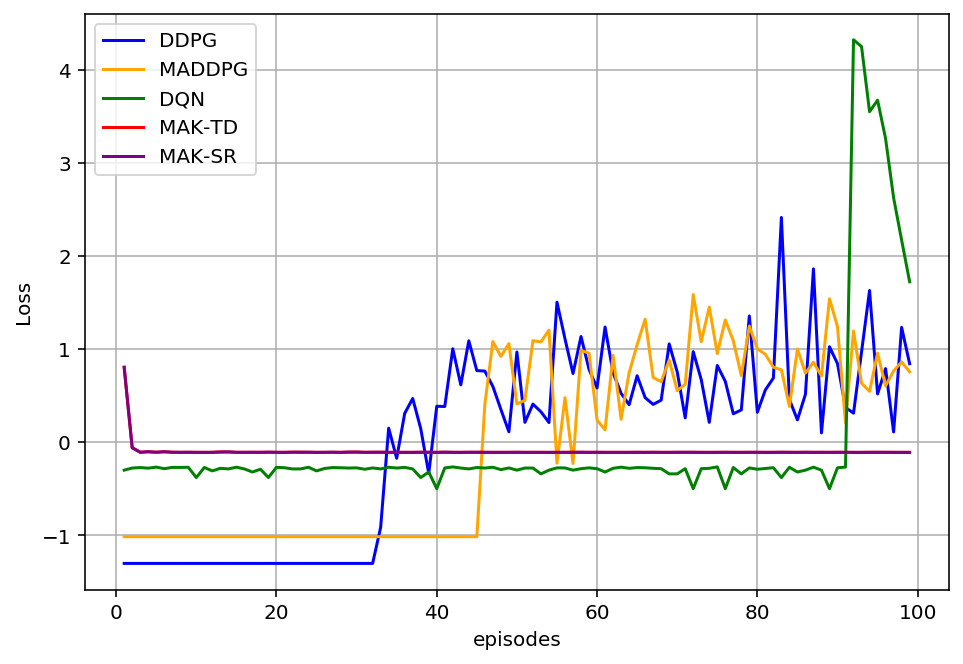}
    \caption{}
  \end{subfigure}
  \begin{subfigure}{6cm}
    \centering
    \includegraphics[width=6.3cm]{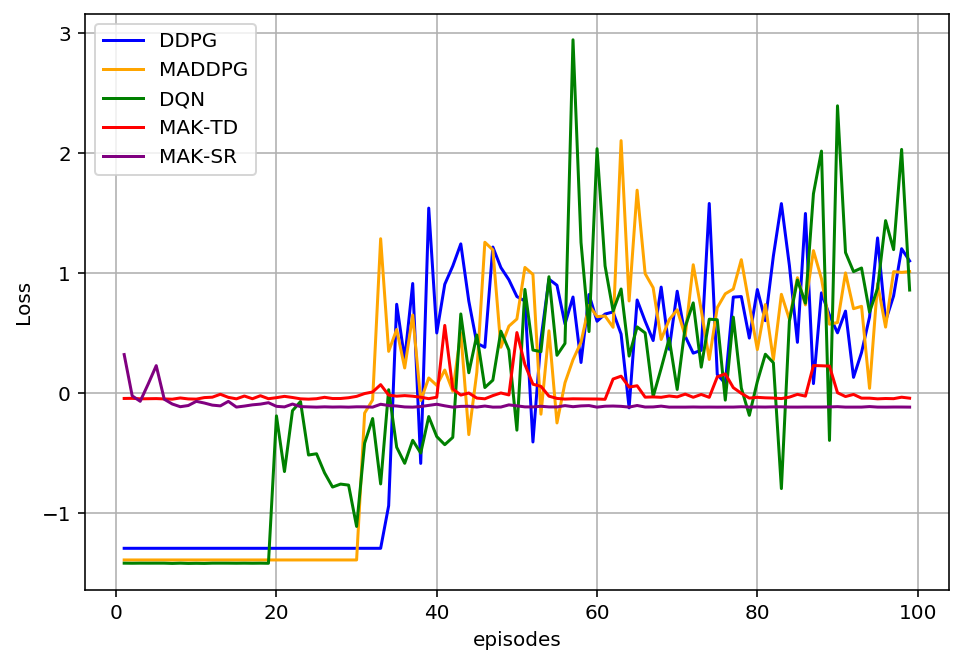}
    \caption{}
  \end{subfigure}
 \\
  \begin{subfigure}{6cm}
    \centering
    \includegraphics[width=6.3cm]{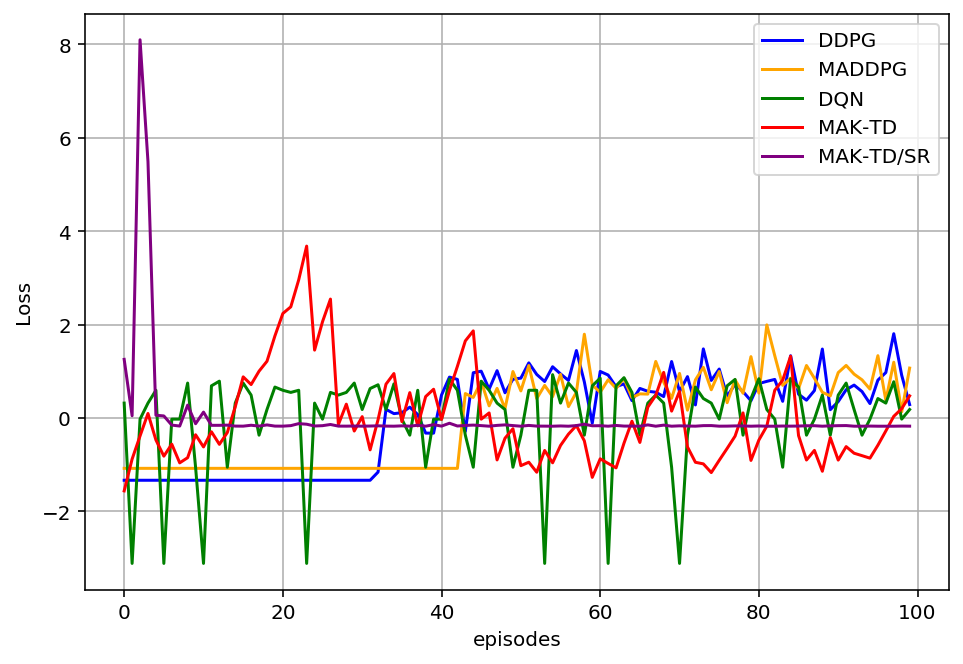}
    \caption{}
  \end{subfigure}
  \begin{subfigure}{6cm}
    \centering
    \includegraphics[width=6.3cm]{loss_1v2.png}
    \caption{}
  \end{subfigure}
\caption{\footnotesize Four different normalized loss functions results for all the agents in the for the four algorithms in four different environments (a) Simple Cooperation  (b) Simple Competition (c) Predator-Prey 2v1, and (d) Predator-Prey 1v2}\label{Fig:18}
\end{figure}

\begin{figure}[t!]
\centering
  \begin{subfigure}{6cm}
    \centering
    \includegraphics[width=6.3cm]{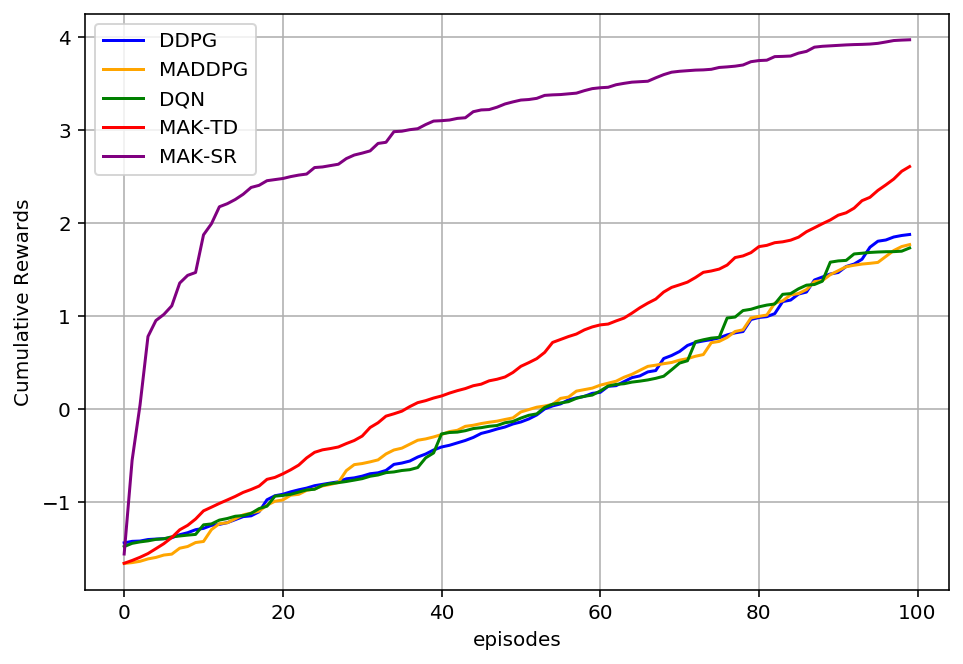}
    \caption{}
  \end{subfigure}
  \begin{subfigure}{6cm}
    \centering
    \includegraphics[width=6.3cm]{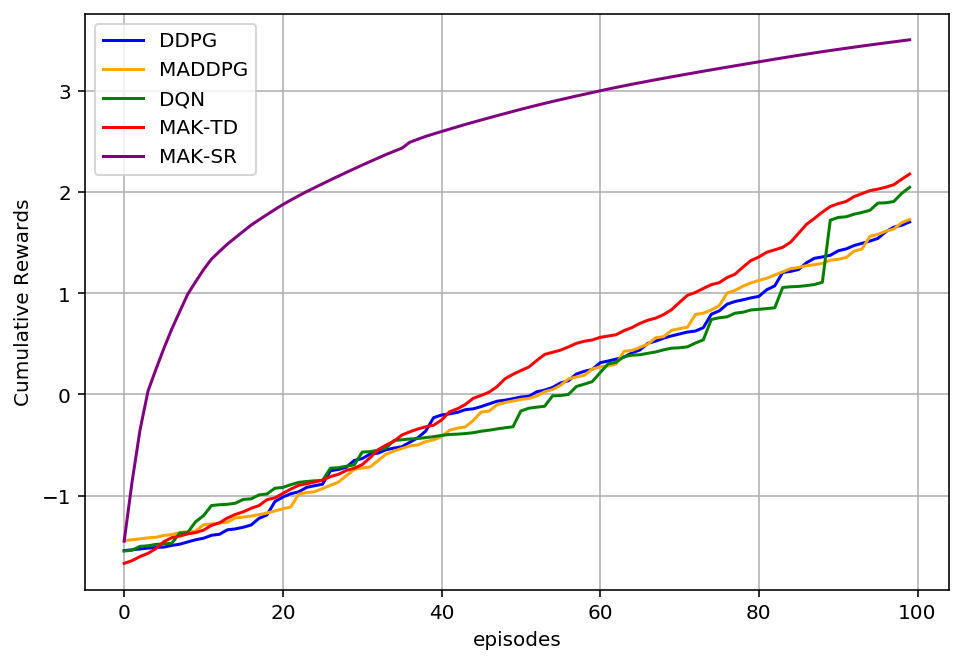}
    \caption{}
  \end{subfigure}
 \\
  \begin{subfigure}{6cm}
    \centering
    \includegraphics[width=6.3cm]{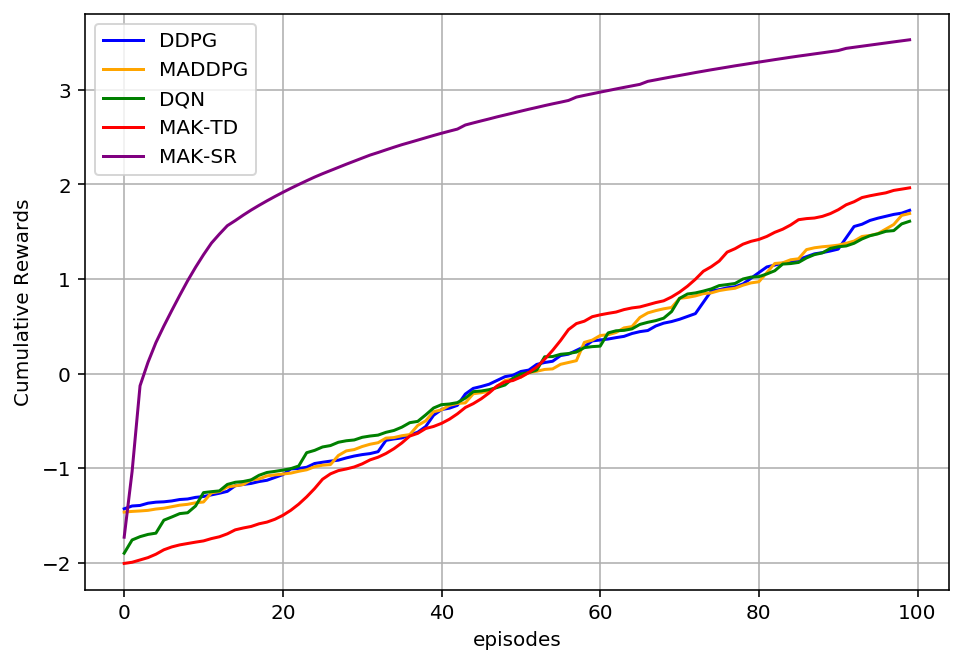}
    \caption{}
  \end{subfigure}
  \begin{subfigure}{6cm}
    \centering
    \includegraphics[width=6.3cm]{rewards_1v2.png}
    \caption{}
  \end{subfigure}
\caption{\footnotesize Four different Reward functions results for all the agents in the for the five algorithms in four different environments (a) Simple Cooperation  (b) Simple Competition (c) Predator-Prey 2v1, and (d) Predator-Prey 1v2}\label{Fig:20}
\end{figure}

\subsection{Results}  \label{sec:Results}
Initially, the agents are trained over different number of episodes, after which $10$ iteration each of  $1,000$ episodes is implemented for testing to compute different results evaluating performance and  efficiency of the proposed $\MSR$ frameworks. First, to evaluate stability of the incorporated RBFs, a Monte-Carlo (MC) study is conducted where $10$ RBFs are used across all the environments. The results are averaged over multiple realizations leveraging MC sampling as shown in Tables~\ref{Table:table1}-\ref{Table:table4}. The results illustrate the inherent stability of the utilized RBFs and the proposed $\MMK$ and $\SR$ frameworks. Capitalizing on the results of Tables~\ref{Table:table1}-\ref{Table:table4}, the $\SR$ can be considered as the most sample efficient approach. It is worth noting that although $\SR$ outperforms the $\MMK$ approach, we included both since the learned representation is not transferrable between optimal policies in the SR learning. For such scenarios, $\MMK$ is an alternative solution providing, more or less, similar performance to that of the $\SR$. To be more precise, when solving a previously unseen MDP, a learned SR representation can only be used for initialization. In other words, the agents still have to adjust the SR representation to the policy, which is only optimal within the existing MDP. This limitation urges us to represent the $\MMK$ as another trusted solution.

\begin{figure}[t!]
\centering
  \begin{subfigure}{6cm}
    \centering
    \includegraphics[width=6.3cm]{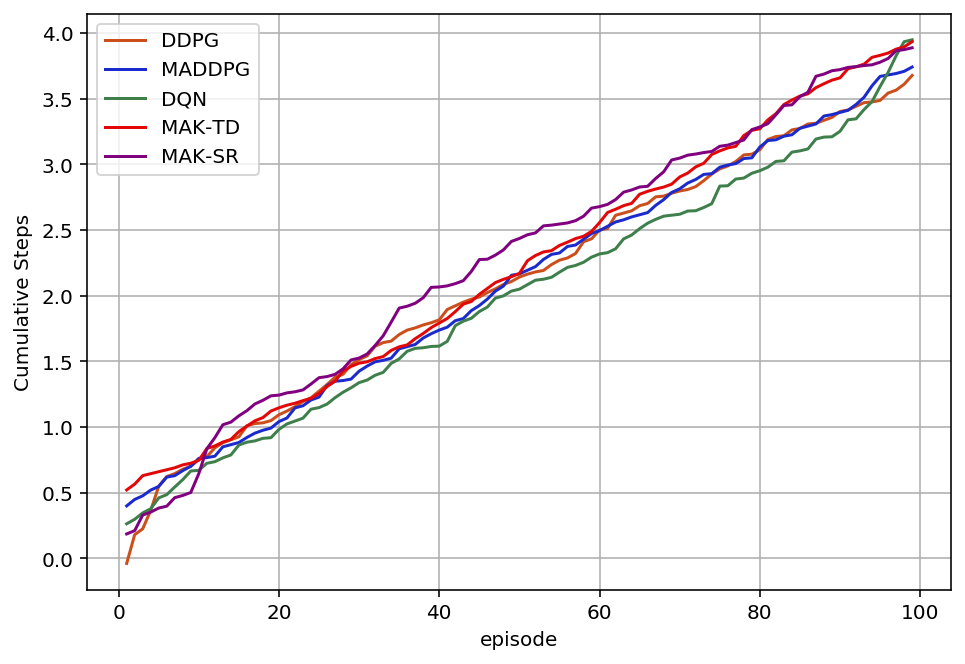}
    \caption{}
  \end{subfigure}
  \begin{subfigure}{6cm}
    \centering
    \includegraphics[width=6.3cm]{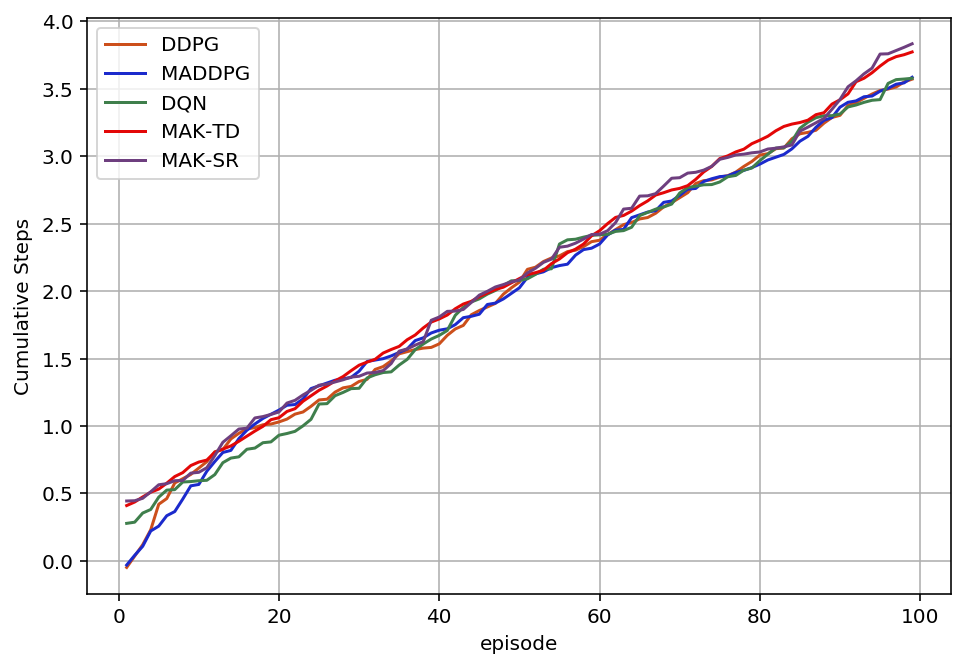}
    \caption{}
  \end{subfigure}
 \\
  \begin{subfigure}{6cm}
    \centering
    \includegraphics[width=6.3cm]{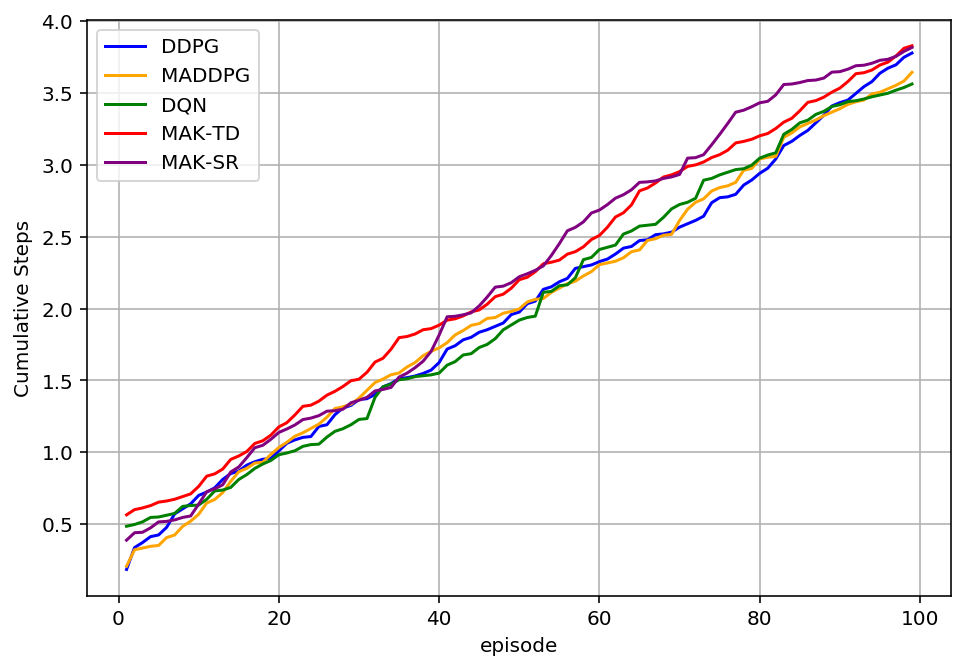}
    \caption{}
  \end{subfigure}
  \begin{subfigure}{6cm}
    \centering
    \includegraphics[width=6.3cm]{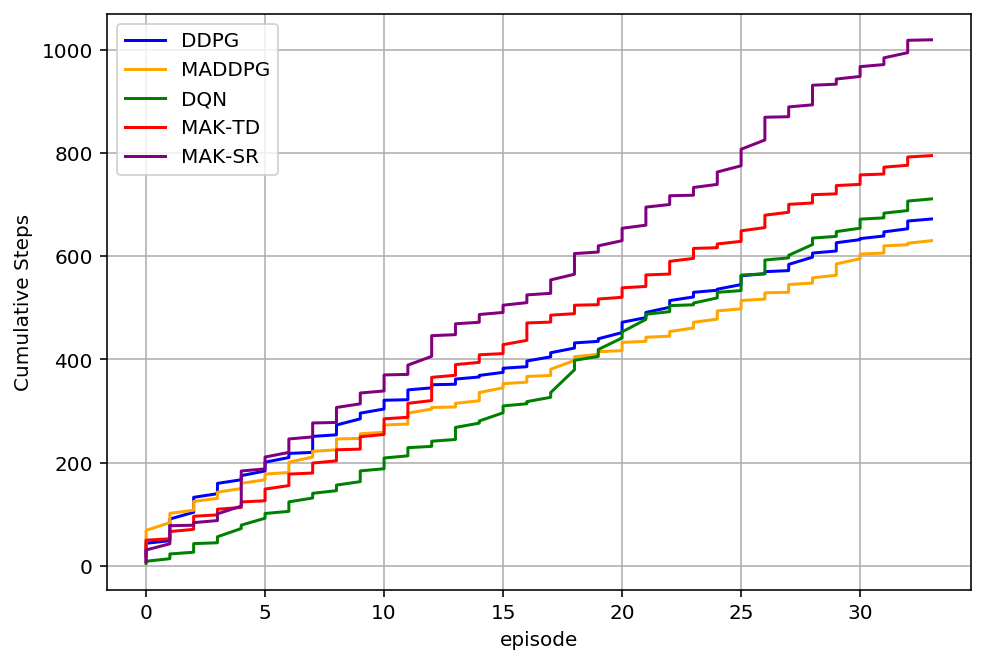}
    \caption{}
  \end{subfigure}
\caption{\footnotesize Number of steps taken in four different environments based on the five implemented algorithms (a) Simple Cooperation  (b) Simple Competition (c) Predator-Prey 2v1, and (d) Predator-Prey 1v2}\label{Fig:23}
\end{figure}

As it can be seen from Table~\ref{Table:table1}, the average loss associated with the proposed $\SR$ is better than that of the $\MMK$. Both frameworks, however, outperform their counterparts, which can be attributed to their improved sample selection efficiency. This excellence can also be seen for the Predator-Prey $1v2$ environment in Fig.~\ref{Fig:fig3}(a). The calculated losses mostly have small values after the beginning of the experiments, indicating stability of the implemented frameworks. As can be seen, other approaches cannot provide that level of performance that is achieved by $\SR$ and $\MMK$ with such low number of training episodes in this experiment. The other three DNN-based approaches can reach such an efficiency with a much higher amount of experience (more than $10,000$ experiments) and use much more memory space to save the batches of the information. Fig.~\ref{Fig:fig3}(b) shows the rewards gained by all the agents in a Predator-Prey environment. As can be seen in Table~\ref{Table:table3} and Fig.~\ref{Fig:fig3}(b), the rewards gained in the $\SR$ are also better than those of the $\MMK$ and are much higher than the other approaches. This can be considered exceptional considering the limited utilized experience. For all other environments this better performance in the gained reward can be seen in Fig.~\ref{Fig:20} where four different reward functions for five discussed algorithms in four experiment environments are shown. It is worth mentioning that the average number of the steps taken by all the agents in the defined environments are also represented in Table~\ref{Table:table4}, showing $\SR$ remarkable results in contrast with the other algorithms. Results related to the number of the steps are also shown in Fig.~\ref{Fig:23} for different environments admitting superiority of the $\SR$ framework in contrast with other solutions. The loss function associated with each of the five implemented methods is shown in Fig.~\ref{Fig:18}. As expected, the performance of each model improves over time as being trained through different training episodes. The proposed $\SR$ and $\MMK$ provide exceptional performances given the small number of training episodes utilized in these experiments. $MADDPG$, $DDPG$, and  $DQN$, however, fail to achieve the same performance level.

\section{Conclusion}  \label{sec:con}
In this paper, an adaptive Kalman filter-based framework, referred to as the Multi-Agent Adaptive Kalman Temporal Difference  ($\MMK$), and its Successor Representation (SR)-based variant (the $\SR$) are introduced as an efficient alternative to Deep Neural Network (DNN)-based Multi-Agent Reinforcement Learning (MARL) solutions. The main objective is to address sample inefficiency, memory problems, and lack of prior information issues of DNN-based MARL techniques. Intuitively speaking, we capitalize on unique characteristics of Kalman Filtering (KF) such as uncertainty modeling and  online second order learning. The proposed $\MSR$ frameworks consider the continuous nature of the action-space that is associated with high dimensional multi-agent environments and exploit Kalman Temporal Difference (KTD) to address the parameter uncertainty. By leveraging the KTD framework, SR learning procedure is modeled into a filtering problem, where Radial Basis Function (RBF) estimators are used to  encode the continuous space into feature vectors. On the other hand, for learning localized reward functions, we resort to Multiple Model Adaptive Estimation (MMAE), as a remedy to deal with the lack of prior knowledge on observation noise covariance and observation mapping function. The proposed $\MSR$ frameworks are evaluated via several experiments, which are implemented through the OpenAI Gym multi-agent RL benchmarks. In these experiments, different number of agents in cooperative, competitive and mixed (cooperative-competitive) scenarios are utilized. The experimental results are discussed illustrating  superior performance of the proposed $\MSR$ frameworks compared to their counterparts. As a final note, the $\SR$ framework outperforms the $\MMK$ approach, however, as the learned representation is not transferrable between optimal policies in the SR learning, $\MMK$ provides an alternative solution with, more or less, similar performance to that of the $\SR$.

\bibliographystyle{IEEEbib}

\end{document}